\documentclass[runningheads]{llncs}
\usepackage[T1]{fontenc}
\usepackage{amsmath}
\usepackage{subfig}
\usepackage{graphicx}
\usepackage{booktabs}
\usepackage{amsmath}
\usepackage{xcolor}
\usepackage{tikz}
\usepackage[export]{adjustbox} 
\begin{document}
\title{Splat-Portrait: Generalizing Talking Heads with Gaussian Splatting}
%\titlerunning{Abbreviated paper title}
% If the paper title is too long for the running head, you can set
% an abbreviated paper title here
\author{Tong Shi\inst{1}\orcidID{0000-0001-5913-4095} \and
Melonie de Almeida\inst{1}\orcidID{0009-0007-4851-0949} \and
Daniela Ivanova\inst{1}\orcidID{0000-0002-3710-7413} \and
Nicolas Pugeault\inst{1}\orcidID{0000-0002-3455-6280} \and 
Paul Henderson\inst{1}\orcidID{0000-0002-5198-7445}}
\authorrunning{T.~Shi et al.}
% First names are abbreviated in the running head.
% If there are more than two authors, 'et al.' is used.
\institute{School of Computing Science, University of Glasgow \\
\email{2431206s@student.gla.ac.uk}\\
% \url{http://www.springer.com/gp/computer-science/lncs}\\
% \email{\{abc,lncs\}@uni-heidelberg.de}
}
\maketitle % typeset the header of the contribution

\begin{abstract}
% The abstract should briefly summarize the contents of the paper in
% 150--250 words.
Talking Head Generation aims at synthesizing natural-looking talking videos from speech and a single portrait image. Previous 3D talking head generation methods have relied on domain-specific heuristics such as warping-based facial motion representation priors to animate talking motions, yet still produce inaccurate 3D avatar reconstructions, thus undermining the realism of generated animations. We introduce Splat-Portrait, a Gaussian-splatting-based method that addresses the challenges of 3D head reconstruction and lip motion synthesis. Our approach automatically learns to disentangle a single portrait image into a static 3D reconstruction represented as static Gaussian Splatting, and a predicted whole-image 2D background. 
%It then generates natural 4D lip motion conditioned on input audio to improve realism from extreme viewpoints, 
It then generates natural lip motion conditioned on input audio, without any motion driven priors. Training is driven purely by 2D reconstruction and score-distillation losses, without 3D supervision nor landmarks. Experimental results demonstrate that Splat-Portrait exhibits superior performance on talking head generation and novel view synthesis, achieving better visual quality compared to previous works. Our project code and supplementary documents are public available at https://github.com/stonewalking/Splat-portrait.
\end{abstract}

\section{Introduction}
Talking Head Generation (THG) aims to synthesize natural-looking talking videos from conditioning information such as driving speech \cite{zhang2023sadtalker,he2023gaia,ye2024real3d} or driving videos \cite{li2023one,ye2024real3d,ma2023otavatar}. The generation of talking heads has received increasing attention due to its importance in various applications, including digital humans \cite{liu2009analysis}, virtual video conferencing \cite{ye2022perceivingli2023one} and visual dubbing \cite{saunders2024dubbing}. Recent 2D methods \cite{zhang2023sadtalker,xu2024hallo,liu2024vqtalker} have achieved significant improvements in video quality and achieve expressive animation results. However, such 2D methods struggle to generate views with large head pose variations, and are not guaranteed to output 3D-consistent renderings from different poses.

3D THG methods~\cite{ye2024real3d,li2023one,ma2023otavatar,ma2023otavatar} have attracted increasing attention in the past two years, since they simultaneously reconstruct accurate 3D geometry and generate expressive facial motions, allowing realistic and 3D-consistent portrait rendering from arbitrary user-controllable viewpoints. The majority of such methods focus on personal talking head generation \cite{peng2024synctalk,yu2024gaussiantalker}, where they overfit a single person's head; they maintain realistic 3D geometry and preserve rich texture details. However, in this work, we consider the more challenging setting where we synthesise a 3D talking head given just a single 2D image, learning a model that generalizes across identities even without 3D supervision.

To represent 3D or 4D faces in THG, Neural Radiance Fields (NeRF) \cite{mildenhall2021nerf} or 3D Gaussian Splatting (3DGS)~\cite{kerbl20233d} are commonly used. NeRF-based methods often exhibit problems such as visual jitters, unsynchronized lip movements, and rendering artifacts; this is because the implicit definition of NeRF entangles static facial geometry with dynamic motion, complicating simultaneous control of lip motions and 3D geometry reconstruction.

Other works \cite{taubner2024cap4d,aneja2024gaussianspeech,chu2024gagavatar} have explored 3D Gaussian Splatting (3DGS)~\cite{kerbl20233d} for 3D avatar generation in the person-generic setting. Compared to NeRF, 3DGS not only improves inference speed and visual quality, but is also more controllable due to its explicit point-cloud-based representation; this makes it possible to animate facial movements more directly and intuitively. For example, \cite{yu2024gaussiantalker} drives Gaussian point clouds for facial motion using parametric 3D facial models \cite{li2017learning}, but it is still challenging to generalize to a person-generic setting. In addition, significant efforts have been made to design and improve 4D animation conditioned on driving information \cite{ye2024real3d,aneja2024gaussianspeech,yu2024gaussiantalker}; the model reconstructs 3D geometry from a single portrait image and learns the corresponding facial motions. These motions are predicted from either an audio sequence or a video sequence driven by motion representation priors, e.g., PNCC, SECC and FLAME~\cite{zhu2016face,li2023one,li2017learning}. These methods relax the difficulty of model training by injecting domain priors, e.g.~3D distillation \cite{chan2022efficient} and motion driven priors, which can lead to unnatural results with limited 3D texture details.

Overall, existing works either reconstruct accurate 3D geometry but require multi-view inputs; or they learn it from monocular videos but yield inaccurate geometry. 

In this work, we introduce Splat-Portrait, a novel audio-driven THG method based on 3DGS (see Fig.~\ref{fig:1}). Our method operates in the single-view setting---it reconstructs the 3D shape of the head directly from one image, outputting pixel-aligned Gaussian splats. To enable lip motion during speech, our model learns to directly animate these splats conditioned on an audio sequence.

Our approach is self-supervised from monocular videos only, and does not rely on 3D morphable models such as FLAME to represent facial shape and expression. We first train our model for static splat reconstruction on a large dataset without audio, then fine-tune on a smaller dataset of portrait videos to learn the correct splat dynamics. This strategy avoids 3D supervision, with the exception of easily-obtained approximate camera intrinsics and extrinsics. During the fine-tuning stage, to further improve the realism of extreme viewpoints that are rare in the training data, we adopt score distillation sampling (SDS) \cite{poole2022dreamfusion}, to extract knowledge from a powerful 2D diffusion prior \cite{karras2022elucidating}. 

Existing works \cite{li2024generalizable,li2023one} typically model only the head region, or model the head and torso regions as a whole, while disregarding the background. This results in a video of a `floating head', rather than a realistic video of the talking head in context. To address this, our model also predict a static RGB background image, and alpha-blend the rasterized splats over this. Driven only by the unsupervised frame-prediction loss, our model automatically learns to reduce the opacity of splats in the background region, and to inpaint the background even behind the head, resulting in realistic disocclusions when the head rotates.

In summary, our main contributions are as follows:
\begin{itemize}
    \item A novel model architecture that disentangles a single portrait image into an accurate 3D splat representation of the head over an inpainted 2D background.
    \item Given audio sequences and corresponding time deltas, we show how to directly animate the 3D splats by predicting and adding dynamic offsets, without any complex motion representation such as a deformation model.
    \item A self-supervised training recipe that uses only monocular videos, without 3D supervision, and integrates knowledge from a strong 2D face prior, distilling its knowledge to improve reconstruction of extreme views.
\end{itemize}
Experimental results on the HDTF \cite{zhang2021flow} and TH-1KH \cite{wang2021one} datasets demonstrate that our approach yields higher video fidelity and quality compared with OTAvatar \cite{ma2023otavatar}, HiDe-NeRF \cite{li2023one}, Real3D-Portrait\cite{ye2024real3d}, and NeRFFaceSpeech \cite{kim2024nerffacespeech} and GAGavatar\cite{chu2024gagavatar}.

\section{Related Work}
\label{sec:related work}
\subsection{3D Head Reconstruction.}
3D Gaussian Splatting (3DGS) \cite{kerbl20233d} has emerged as a popular method for 3D head reconstruction due to its efficient rendering speed and superior reconstruction quality \cite{yu2024gaussiantalker,taubner2024cap4d,wang2023gaussianhead,aneja2024gaussianspeech}. Neural Radiance Fields (NeRF) \cite{mildenhall2021nerf} have been widely adopted for 3D talking head generation; NeRF represents scenes through volumetric radiance fields encoded by neural networks, enabling photorealistic renderings from novel viewpoints. NeRF-based methods have naturally extended into talking-head synthesis \cite{peng2024synctalk,guo2021ad}. Early NeRF-driven approaches for talking-head reconstruction \cite{ma2023otavatar,guo2021ad} often require subject-specific training, limiting their scalability. Recent methods leverage 3DGS to address these limitations by significantly improving rendering speed and depth estimation \cite{kerbl20233d,yu2024gaussiantalker}. 3DGS represents scenes explicitly with discrete geometric primitives (3D Gaussians), enabling efficient optimization and real-time rendering. Notably, Rivero et al. \cite{rivero2024rig3dgs} introduced a dynamic head reconstruction framework using 3DGS, and GaussianHead \cite{wang2023gaussianhead} further advanced these capabilities. By binding the Gaussians to an underlying geometric model, dynamic talking heads can be generated. However, these works for directly regressing 3D representations require prediction in a canonical space, which often fails to handle extreme head poses or significant appearance variations, such as non-photorealistic or animated scenarios. Current techniques still exhibit overfitting issues and rely heavily on domain priors during training, such as the parametric FLAME model \cite{li2017learning}. Our method builds upon 3DGS to reconstruct dynamic talking heads directly from a single image.

\subsection{Probabilistic 3D Reconstruction.}
Single-view 3D head reconstruction~\cite{dhamo2024headgas} is an ambiguous problem due to the fact that training data usually have limited variation in poses, particularly in face monocular videos. Recently, diffusion models have been employed for conditional novel view synthesis~\cite{watson2022novel} and also multi-view synthesis~\cite{taubner2024cap4d}. Since the results usually have ambiguous geometry, the output rendered results can exhibit noticeable artifacts, particularly a lack of texture details in unseen views. This can be mitigated by distilling prior knowledge from a 2D model~\cite{poole2022dreamfusion}. Existing 3D reconstruction works found that distilling knowledge from 2D images could help to make the 3D representation much more controllable by reconstructing a geometry at every step of the denoising process~\cite{tewari2023diffusion}. Other works pre-train a robust reconstructor~\cite{liu2023one} and use a 3D prior~\cite{muller2023diffrf} which can be used in an image-conditioned auto-decoding framework. However, their work is complex and computationally heavy to train. We also leverage a pretrained 2D generative prior when training for 3D reconstruction; this helps our method with extreme-view 3D head reconstruction, but avoids expensive iterative sampling.

\subsection{Face Animation.}
Initial efforts for talking head animation utilized 2D approaches, employing generative adversarial networks, image-to-image translation \cite{isola2017image} or diffusion models \cite{stypulkowski2024diffused}, to generate facial animations. Most 2D talking head generation methods design a mapping relationship between face images and audio feature. These methods \cite{zhang2023sadtalker,xu2024hallo} often underestimate detailed individual differences. Recently, 3D facial animation methods \cite{li2023one,ye2024real3d} became popular, however they adopt PNCC SECC as driving features, leading to unnatural expressions and lip motion. Some warping-based methods \cite{thies2016face2face,ye2024real3d} employ 3DMMs, or face blend shapes, which support animation via disentangled representation of shape, expression and pose. However, these approaches can fall short of accurately reproducing a talking face due to limited amplitude, leading to shortcomings in identity preservation and pose controllability. Our approach is designed to directly edit the 3D representation to animate lip motion over time.

\section{Methodology}
\label{sec:methods}
\begin{figure*}
  \centering
  \resizebox{1\linewidth}{!}{\includegraphics{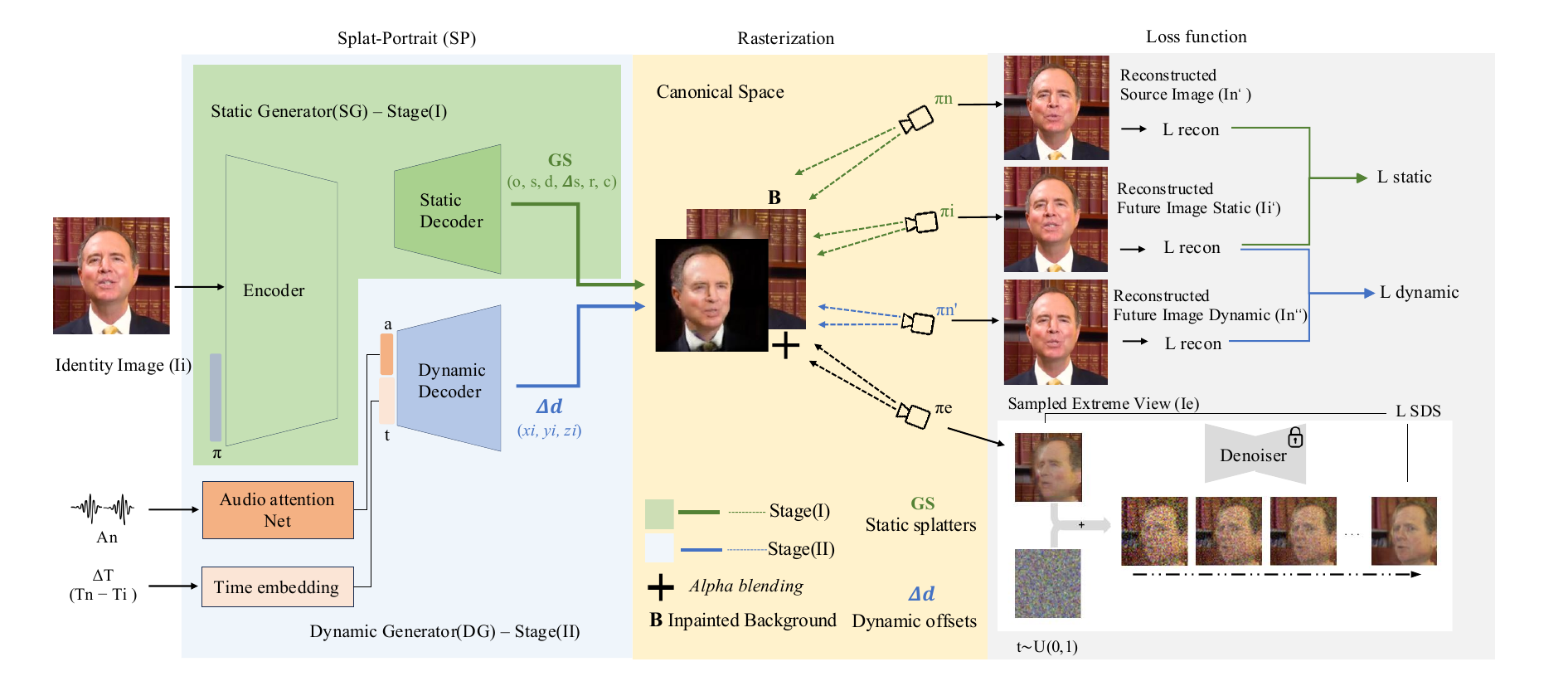}}
  \vspace{-10pt}
  \caption{Overview of Splat-Portrait. The identity image \( I_i \) is passed through a U-Net Static Generator(SG) to reconstruct static 3D Gaussian Splats, alpha-blended over a predicted 2D background. The dynamic decoder estimates splat offsets at timestep \( T_n \) using audio features \( A_n \) and time embedding \( \Delta T \). The training procedure consists of two stages, stage(I): an initial pre-training phase, where the static components are trained on a large-scale dataset using a static reconstruction loss \( \mathcal{L}_{\mathrm{static}} \), and stage(II): a fine-tuning phase on a smaller dataset incorporating an additional dynamic reconstruction loss \( \mathcal{L}_{\mathrm{dynamic}} \). And a score distillation loss $\mathcal{L}_{\mathrm{SDS}}$ on extreme viewpoints applied during both stages.
  % \( \pi_e \), cropping and aligning them before input into a 2D diffusion prior. We compute the SDS loss between synthesized images \( I_e' \) and target images \( I_e \).
  }
  \label{fig:1}
\end{figure*}
The overall architecture of our method Splat-Portrait is illustrated in Fig.~\ref{fig:1}. Splat-Portrait consists of two main stages: (1) pre-training to reconstruct 3D static splats (Sec.~\ref{subsec:static-generator}); (2) fine-tuning with an audio-conditioned dynamic decoder (Sec.~\ref{subsec:portrait-generator}), while also using score distillation (Sec.~\ref{subsec:sds}) to refine appearance from extreme viewpoints. 

\subsection{Static Splat Generation}
\label{subsec:static-generator}
3D Gaussian Splatting (3DGS) \cite{kerbl20233d} uses anisotropic 3D Gaussians as geometric primitives to explicitly represent 3D scenes. For our 3D head reconstruction, we first pre-train a static generator (SG) that outputs pixel-aligned splats, as shown in Fig.~\ref{fig:1}. The design of SG is based on Splatter-Image~\cite{szymanowicz2024splatter}. However, unlike \cite{szymanowicz2024splatter} we do not have access to wide-baseline multi-view images for training; instead we use more challenging monocular video data.
We also predict an inpainted 2D RGB background as well the per-pixel 3D splat attributes, and alpha-blend the rasterized splats over this. 
During training, we randomly choose pairs of frames from a video, denoted source image $I_i$ and future image $I_n$. Given \( I_i \), the network predicts a set of Gaussian Splatting parameters $GS$, at each pixel: opacity \(o\), scale \(s\), depth \(d\), static offset \(\Delta_s\), rotation \(r\), splat colour \(c\) (encoding per-pixel 3D Gaussian attributes), and the 2D background colour \(RGB\). The view-space 3D position $p$ of the Gaussian at a pixel with ray direction $\mathbf{r}$ is then given by $p = \mathbf{r}\, d + \Delta_s$.

During training, we feed the network with \(I_i\) at time step \( Ti \). Additionally, we inject the approximate camera-to-world translation and focal length $\pi$. We do so by encoding each entry via a sinusoidal positional embedding of order 9, resulting in 60 dimensions in total. These are applied to the U-Net blocks via FiLM \cite{perez2018film} conditioning. During our experiments, we found this helps with convergence of depth predictions.

Given the Gaussian attributes described above, we use the differentiable rasterizer $\mathcal{R}$ from \cite{kerbl20233d} to render the splats at canonical space with static offset enabled to reconstruct images $I_i^*$ and $I_n^*$ at the camera poses of both \(I_i\) and \( I_n \) respectively. We compute a combined L2 and LPIPS reconstruction loss \( \mathcal{L}^{\mathrm{rec}}_{\mathrm{static}} \) between corresponding rendered and ground-truth images, i.e.
\begin{equation}
\mathcal{L}{\mathrm{static}} = 
||I_i - I_i^* ||_2 +
||I_n - I_n^* ||_2 +
\lambda_{\text{LPIPS}} \left[
\mathcal{L}_\text{LPIPS}(I_i,\, I_i^*) +
\mathcal{L}_\text{LPIPS}(I_n,\, I_n^*)
\right]
\end{equation}
Here the LPIPS term combines VGGface and VGG19 features, and the weight $\lambda$ is empirically set to 0.01.
For each image, we render (and calculate the loss) twice, once with a random coloured background and once with our predicted 2D background alpha-blended behind the splats. We found that this helps to improve the colour and opacity for both background and foreground regions, without incorporating any mask supervision. 

\subsection{Audio-Conditioned Dynamic Splats}
\label{subsec:portrait-generator}
For predicting audio-conditioned dynamics representing lip movements, we designed a dynamic decoder with skip connections from the SG decoder.
We only use this during the fine-tuning stage, after a good static reconstruction model has been learnt during pre-training. It predicts time-dependent offsets for every splat, conditioned on the audio signal and a time delta indicating what instant in the audio we want the splat offsets for. In our experiments we found that including this time delta improves convergence of the dynamic decoder.

For a given input frame \(I_i\), and future frame \( I_n \) plus its contemporaneous audio segment, we first extract audio features using \texttt{Wav2Vec2-XLSR 53}~\cite{conneau2020unsupervised}. 

Our model employs dedicated networks to fuse audio and temporal information effectively. Specifically, audio features are first encoded through an audio feature extraction module (AudioNet), which comprises several 1D convolutional layers followed by fully connected layers to yield compact audio embeddings. These embeddings are further refined through an attention-based network (AudioAttNet); this consists of a series of convolutional layers with decreasing channel sizes (from 16 to 1) interleaved with LeakyReLU activations. The output from these convolutional layers is then reshaped and passed through a linear layer followed by a softmax operation to calculate attention weights across the audio sequence. The weighted audio embeddings are summed to produce a refined audio representation capturing temporal dependencies across audio frames. For temporal embeddings, positional encoding or Fourier-based embeddings are utilized to encode timestep information; then audio and temporal embeddings are combined to form the conditioning feature. This combined embedding is injected into the dynamic decoder using FiLM conditioning, allowing the audio and time delta to control the generated motion.

Our dynamic decoder outputs a dynamic offset \(\Delta_d\) for the splat at each pixel, conditioned on time $T$. Hence the splat position at time $T$ is $p_T = p + \Delta_d$.
To effectively train our model and maintain the static reconstruction ability learnt during pre-training, we adopt both \( \mathcal{L}_{\mathrm{static}} \) loss and the SDS loss introduced in Sec.~\ref{subsec:sds}. We render the source frame $I_i^*$ with dynamic offsets fixed to zero (as in Sec.~\ref{subsec:static-generator}), but now render the future frame $I_n^{**}$ using the predicted offsets. Our dynamic reconstruction loss in the fine-tuning stage is: 
\begin{equation}
\mathcal{L}{\mathrm{dynamic}} = 
||I_i - I_i^* ||_2 +
||I_n - I_n^{**} ||_2 +
\lambda_{\text{LPIPS}} \left[
\mathcal{L}_\text{LPIPS}(I_i,\, I_i^*) +
\mathcal{L}_\text{LPIPS}(I_n,\, I_n^{**})
\right]
\end{equation}

\subsection{Distillation from a 2D diffusion prior}
\label{subsec:sds}
In the fine-tuning stage, we also use score distillation \cite{poole2022dreamfusion} to extract knowledge from a 2D diffusion model \cite{karras2022elucidating} to improve the appearance of extreme poses. We first render our predicted reconstruction at a randomly sampled extreme pose, then crop and align the image following \cite{karras2019style} to match the distribution learnt by the 2D diffusion model. Given this aligned image \( x_{\text{clean}} \), we then add a random amount of noise then run the reverse diffusion process.

Specifically, we define a sequence of noise levels \(\sigma\) as follows:
\begin{equation}
\sigma_i = \left[\sigma_{\text{max}}^{\frac{1}{\rho}} + \frac{i}{N - 1}\left(\sigma_{\text{min}}^{\frac{1}{\rho}} - \sigma_{\text{max}}^{\frac{1}{\rho}}\right)\right]^\rho,
\end{equation}
where \( \sigma_{\text{max}} \) and \( \sigma_{\text{min}} \) denote maximum and minimum noise levels, \( \rho \) is a hyper parameter controlling the distribution of timesteps, and \(N\) is the total number of discretized steps. We choose the noise level from 60\%--80\% of the original range used in training the diffusion model, since we found this range effectively preserves the portrait’s overall appearance while significantly improving texture inpainting for extreme viewpoints.

The noised image at the initial timestep \( t_0 \) is generated by adding Gaussian noise to the normalized input image, i.e.~$
x_{\text{noised}} = x_\text{clean} + \sigma_0 \cdot \epsilon$, where $\epsilon \sim \mathcal{N}(0, I)$.
For each subsequent timestep, we perform an Euler integration step to progressively denoise the image. Specifically, given the current timestep \( t_{\text{cur}} \) and next timestep \( t_{\text{next}} \), the Euler step is computed as:
\begin{equation}
x_{\text{next}} = x_{\text{cur}} + (t_{\text{next}} - t_{\text{cur}}) \cdot d_{\text{cur}}, \quad d_{\text{cur}} = \frac{x_{\text{cur}} - \text{net}(x_{\text{cur}}, t_{\text{cur}})}{t_{\text{cur}}},
\end{equation}
where $\text{net}$ represents the pre-trained denoiser model.

This sampling procedure yields a denoised face image (the final $x_\text{next}$) that is similar to the original rendered one, but more realistic according to the diffusion prior. We define a loss $\mathcal{L}_\text{SDS}$ as the L2 reconstruction loss between the rendered frame \( x_{\text{clean}} \) and the denoised frame, back-propagating only into the former. This guides our rendered frames to look more like similar realistic samples from the diffusion model. During training, we randomly sample extreme viewpoints following a bullet-effect trajectory, with pitch variations up to $\pm 12.5^{\circ}$ and yaw variations up to $\pm 45^{\circ}$ from the canonical view, and apply the SDS loss between $I_i$ and $x_{\text{clean}}$. Note that unlike \cite{poole2022dreamfusion} and common practice, we apply the SDS loss during model training, not during inference, meaning the latter remains very fast.

\subsection{Overall Loss}
Our total losses are defined as follows.
For stage one (static pretraining):
\begin{equation}
\mathcal{L}_{\text{total\_static}} =
\mathcal{L}_{\text{static}}(I_i,\, I_n) +
\mathcal{L}_{\text{SDS}}.
\end{equation}
For stage two (audio-conditioned fine-tuning):
\begin{equation}
\mathcal{L}_{\text{total\_dynamic}} =
\mathcal{L}_{\text{dynamic}}(I_i,\, I_n) +
\mathcal{L}_{\text{SDS}}.
\end{equation}
 In both cases, we use AdamW for optimization, with a learning rate of $2.5 \times 10^{-5}$ and weight decay of $10^{-5}$.

\section{Experiments}
\label{sec:experiments}
\paragraph{Datasets and Implementation Details.}
We evaluate our approach on two widely used datasets of monocular talking portrait videos -- HDTF \cite{zhang2021flow} and TalkingHead-1KH \cite{wang2021one}. HDTF consists of over 400 samples of talking videos from over 350 subjects. For TalkingHead-1KH, we manually select 1100 identity videos following a similar distribution as HDTF, such that there is no occlusion over the torso and mouth, and with static background. Each identity video contains minimum 300 frames and maximum 10000 frames. We extract frames at 25Hz, and the audio sampling rate is 16kHz. We resize the image frames to 256 \( \times \) 256. Following the steps in \cite{peng2024synctalk}, we follow \cite{guo2021ad} to use 3DMM optimization to extract approximate intrinsic and extrinsic camera parameters. We use the complete video clips (often with substantial camera motion) for training. For evaluation, we randomly sample 50 identity videos as test sets, and use the first 5s of each. We adopt the SongUNet \cite{song2020denoising} architecture for our static encoder and dynamic decoder.

\paragraph{Metrics.} We measure the quality of synthetic images using structural similarity (SSIM), peak signal-to-noise ratio (PSNR), Learned Perceptual Image Patch Similarity LPIPS~\cite{zhang2018unreasonable}, and Fr\'{e}chet Inception Distance (FID); we use Cosine similarity (CSIM) for measuring identity preservation, and SyncNet \cite{Chung16a} to measure lip synchronization scores (LipSync).

\paragraph{Baselines.}
We compare our approach to several existing 3D talking head generation works. \textbf{OTAvatar}~\cite{ma2023otavatar} is a video-driven method that uses a pre-trained 3D GAN to obtain a 3D talking portrait video; \textbf{HiDe-NeRF}~\cite{li2023one}, a 3D talking face model that uses a motion prior and deformation field for face animation; \textbf{Real3D-Portrait}~\cite{ye2024real3d}, a nerf-based method that uses images generated by EG3D to train a 3D model; \textbf{NeRFFaceSpeech}~\cite{kim2024nerffacespeech} one nerf-based audio driven method for synthesising talking head video, and the state-of-the-art \textbf{GAGavatar}~\cite{chu2024gagavatar}. Additionally in the audio-driven setting, we extend GAGavatar with ARtalker \cite{chu2025artalk}. Note OTAvatar and HiDe-NeRF are video-driven methods, they are not directly driven by audio; for fair comparison, we use the same identity video as driving video for evaluation. We set the input image size as 256$\times256$ to enable fair comparison, upsampling for methods that require this. We compare the baselines using their preferred masking and cropping settings.

\begin{figure}[t]
  \centering
  \resizebox{0.95\linewidth}{!}{\includegraphics{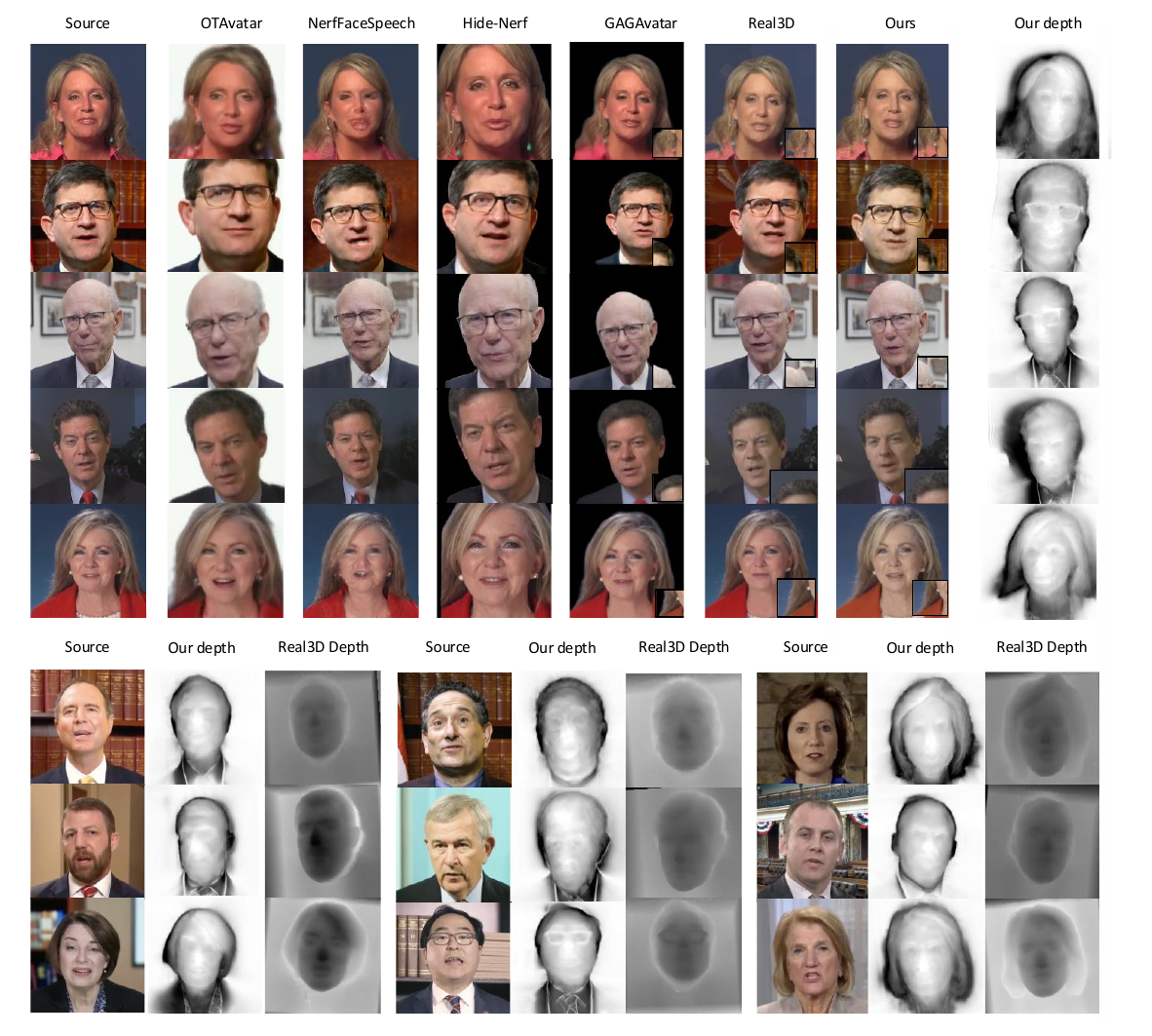}}
  \vspace{8pt}
  \caption{Qualitative results. \textbf{Top:} We show source frames from five videos, future predicted frames from ours and baselines, and future depths from ours.
  \textbf{Bottom:} Additional examples of 3D reconstruction, for our method and Real3D-Portrait, displaying the input frame, and the reconstructed depth-map from each method.}
  \label{fig:short-a}
\end{figure}

\subsection{Quantitative Evaluation}
We compare with the baselines in \textit{same identity} and \textit{cross-identity} settings. During testing, the driving motion condition and head pose are obtained from a reference video. Under the same-identity setting, we use the first frame of the reference video as the source image; otherwise, the source image is of a different identity.
For the cross-identity setting, for Real3D-Portrait, we only compare with its audio-driven setting.
Quantitative results concerning the quality and fidelity for the same-identity setting are listed in Tab.~\ref{tab:same-identity}. These show that our method outperforms other state-of-the-art approaches on almost all fidelity metrics. This is despite our method being trained without 3D supervision, using only a dataset of monocular videos. Splat-Portrait achieves the best overall video quality, as well as higher LipSync score, demonstrating that our 3D deformable model without any motion representation could sync well on lip motions.
Moreover, our model achieves the highest performance on CSIM, meaning it has a strong ability to preserve subject identity in different views.
We also compare the baselines with cross-identity evaluation, where the driving videos are obtained from a reference video, and we use another identity for target. Since there is no ground truth for this setting, we evaluate the results only on CSIM, FID and Lip sync. The results are given in Tab.~\ref{tab:cross-identity}.
We see that our method performs best on FID and CSIM, which indicates our model still yields high video generation quality even in this more challenging setting.

\begin{table}[t]
  \centering
  \begin{tabular}{@{}lcccccc@{}}
    \toprule
    Method & PSNR $\uparrow$ & SSIM $\uparrow$ & LPIPS $\downarrow$ & CSIM $\uparrow$ & FID $\downarrow$ & LipSync $\uparrow$ \\
    \midrule
    OTAvatar  & 13.85 & 0.488 & 0.432 & 0.559 & 78.98 & 5.908\\
    NeRFFaceSpeech & 13.90 & 0.520 & 0.480 & 0.580 & 64.60 & 4.880 \\
    HiDe-NeRF   & 21.44 & 0.685 & 0.221 & 0.716 & 28.63 & 5.552\\
    Real3D-Portrait & 22.40 & 0.758 & 0.191 & 0.761 & 35.69 & \textbf{6.681}\\
    GAGAvatar + ARtalker & 23.08 & 0.786  & 0.182 & 0.753 & 37.89 & 6.580\\
    Ours & \textbf{23.87} & \textbf{0.814} & \textbf{0.128} & \textbf{0.811} & \textbf{25.58}  & 6.328\\
    \bottomrule
  \end{tabular}
  \vspace{8pt}
  \caption{Quantitative evaluation of our method and baselines in the same identity setting.}
  \label{tab:same-identity}
\end{table}

\begin{table}[t]
  \centering
  \begin{tabular}{@{}lccccc@{}}
    \toprule
    Method  & CSIM $\uparrow$ & FID $\downarrow$ & LipSync $\uparrow$\\
    \midrule
    NeRFFaceSpeech & 0.450 & 50.80 & 4.423\\
    OTAvatar  & 0.521 & 79.32 & 5.032 \\
    HiDe-NeRF  & 0.628 & 31.23 & 5.652 \\
    Real3D-Portrait & 0.691 & 40.82 & 6.521 \\
    GAGAvatar + ARtalker  & 0.687 & 35.82 & \textbf{6.503} \\
    Ours & \textbf{0.726} & \textbf{28.62} & 6.218 \\
    \bottomrule
  \end{tabular}
  \vspace{8pt}
  \caption{Quantitative evaluation of our method and baselines in the cross-identity setting.}
  \label{tab:cross-identity}
\end{table}

\subsection{Qualitative evaluation}
In this section we provide visual comparisons of all tested methods (see Figure~\ref{fig:short-a}).
We find that our method preserves face texture details, such as hair and wrinkles well, yielding high-quality novel views. In particular, our method preserves details such as earrings which move during the video. Since we do not require the head to be pre-segmented, our method handles fine details at the silhouette edges well, and effectively blends the rendered portrait over the estimated background. Fig~\ref{fig:short-a} also compares depth-maps rendered by our model with those from Real3D-Portrait, to better visualise the quality of the 3D shape. Compared with Real3D-Portrait, it is clear that our method preserves much more detailed geometry information.

\begin{table}[t]
  \centering
  \begin{tabular}{@{}lccccc@{}}
    \toprule
     Method  & PSNR $\uparrow$ & SSIM $\uparrow$ & LPIPS $\downarrow$\\
     \midrule
     w/o time delta &22.68 &0.768 &0.146 \\
     w/o pre-training &23.30&  0.758& 0.149  \\
     w/o SDS  &23.58  & 0.788 & 0.147 \\
     w/o static offset & 23.30 & 0.791 & 0.145 \\
     only future l2 loss & 23.41 & 0.772 & 0.138 \\
     Full (SP) & \textbf{23.87} & \textbf{0.814} & \textbf{0.128} \\
     \bottomrule
    \end{tabular}
    \vspace{8pt}
  \caption{Ablation study showing the benefit of different components of our model.}
  \label{tab:ablation}
\end{table}

\begin{figure}[t]
\centering
\setlength{\tabcolsep}{0pt}
\renewcommand{\arraystretch}{0}
\begin{tabular}{cccccc}\\
& \includegraphics[scale=0.24]{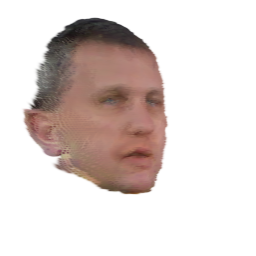} &
  \includegraphics[scale=0.24]{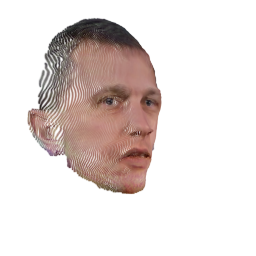} &
  \includegraphics[scale=0.24]{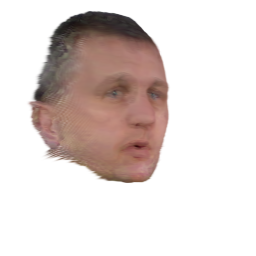} &
  \includegraphics[scale=0.24]{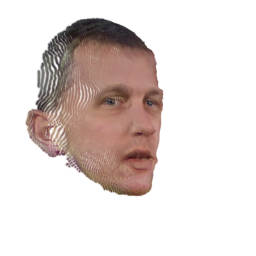} &
  \includegraphics[scale=0.24]{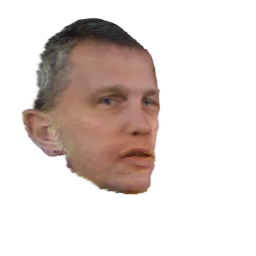} \\
& \includegraphics[scale=0.24]{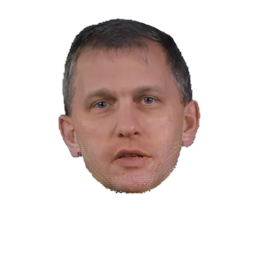} &
  \includegraphics[scale=0.24]{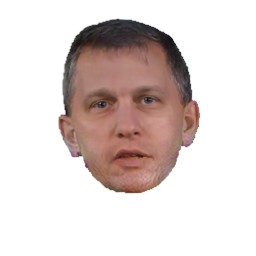} &
  \includegraphics[scale=0.24]{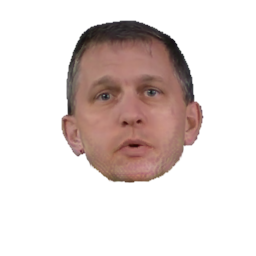} &
  \includegraphics[scale=0.24]{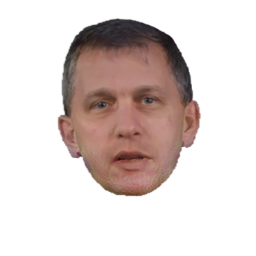} &
  \includegraphics[scale=0.24]{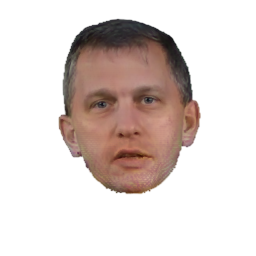} \\
& \includegraphics[scale=0.24]{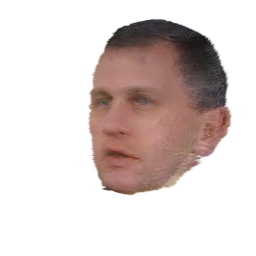} &
  \includegraphics[scale=0.24]{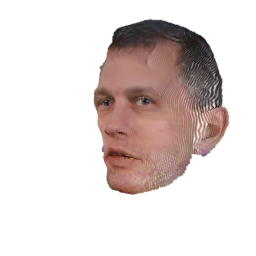} &
  \includegraphics[scale=0.24]{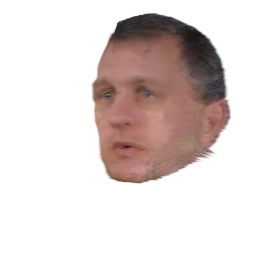} &
  \includegraphics[scale=0.24]{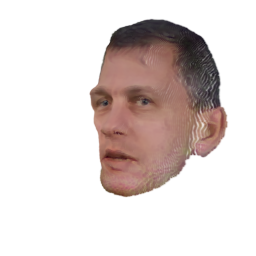} &
  \includegraphics[scale=0.24]{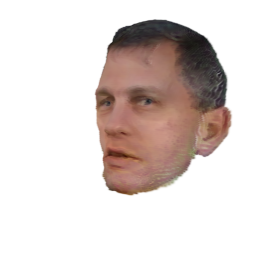} \\
&  w/o pre-training & w/o SDS & w/o static offset & \shortstack[c]{only future\\L2 loss} & Ours
\end{tabular}
\caption{Ablation study, with extreme head yaw angles (top row at -35°, the middle row at 0°, and the bottom row at +35°).}
\label{fig:2}
\end{figure}

\subsection{Ablation Study}
We test four ablations of our model: (1) w/o time delta, which does not inject the time embedding (see Sec.~\ref{subsec:portrait-generator}); (2) w/o pre-training, which does not pre-train the static generator (see Sec.~\ref{subsec:static-generator}); (3) w/o SDS, which omits the score distillation loss during the fine-tuning stage (see Sec.~\ref{subsec:sds}); (4) without static offsets during fine-tuning stage; (5) without the initial-frame reconstruction loss, only the future reconstruction loss. We show the results in Fig.~\ref{fig:2}. Without the pre-training process, we see the 3D geometry accuracy drops significantly, the reconstructed 3D head exhibits flattened geometry, with reduced 3D structure. Training with only one frame for supervision instead of two randomly selected frames, it is hard to reconstruct depths and static offsets (and thus the static shape of the face) well, as some structural information is instead represented in the dynamic offsets. As shown in Fig.~\ref{fig:2}, when enabling static splat offsets, the visualized 3D representation shows a smooth, realistically curved geometry. Lastly our SDS loss greatly enhances the realism of extreme poses.

\section{Conclusion}
We proposed Splat-Portrait for Talking Head Generation. Our method is trained on monocular videos without 3D supervision, yet can synthesize accurate 3D geometry and plausible lip movements directly from a single portrait image, yielding state-of-the-art results. By effectively disentangling static and dynamic attributes and using a score-distillation  loss, Splat-Portrait significantly enhances realism, particularly from extreme viewpoints rarely encountered during training. Additionally, the simplicity and efficiency of our model structure allow it to animate 3D splats effectively without complex deformation models, making it lightweight and practical for real-world applications.

\bibliographystyle{splncs04}
\bibliography{egbib}

@inproceedings{zhang2023sadtalker,
  title={Sadtalker: Learning realistic 3d motion coefficients for stylized audio-driven single image talking face animation},
  author={Zhang, Wenxuan and Cun, Xiaodong and Wang, Xuan and Zhang, Yong and Shen, Xi and Guo, Yu and Shan, Ying and Wang, Fei},
  booktitle={Proceedings of the IEEE/CVF conference on computer vision and pattern recognition},
  pages={8652--8661},
  year={2023}
}

@article{he2023gaia,
  title={Gaia: Zero-shot talking avatar generation},
  author={He, Tianyu and Guo, Junliang and Yu, Runyi and Wang, Yuchi and Zhu, Jialiang and An, Kaikai and Li, Leyi and Tan, Xu and Wang, Chunyu and Hu, Han and others},
  journal={arXiv preprint arXiv:2311.15230},
  year={2023}
}

@inproceedings{peng2024synctalk,
  title={Synctalk: The devil is in the synchronization for talking head synthesis},
  author={Peng, Ziqiao and Hu, Wentao and Shi, Yue and Zhu, Xiangyu and Zhang, Xiaomei and Zhao, Hao and He, Jun and Liu, Hongyan and Fan, Zhaoxin},
  booktitle={Proceedings of the IEEE/CVF Conference on Computer Vision and Pattern Recognition},
  pages={666--676},
  year={2024}
}

@inproceedings{yu2024gaussiantalker,
  title={GaussianTalker: Speaker-specific Talking Head Synthesis via 3D Gaussian Splatting},
  author={Yu, Hongyun and Qu, Zhan and Yu, Qihang and Chen, Jianchuan and Jiang, Zhonghua and Chen, Zhiwen and Zhang, Shengyu and Xu, Jimin and Wu, Fei and Lv, Chengfei and others},
  booktitle={Proceedings of the 32nd ACM International Conference on Multimedia},
  pages={3548--3557},
  year={2024}
}

@article{ye2024real3d,
  title={Real3d-portrait: One-shot realistic 3d talking portrait synthesis},
  author={Ye, Zhenhui and Zhong, Tianyun and Ren, Yi and Yang, Jiaqi and Li, Weichuang and Huang, Jiawei and Jiang, Ziyue and He, Jinzheng and Huang, Rongjie and Liu, Jinglin and others},
  journal={arXiv preprint arXiv:2401.08503},
  year={2024}
}

@inproceedings{stypulkowski2024diffused,
  title={Diffused heads: Diffusion models beat gans on talking-face generation},
  author={Stypu{\l}kowski, Micha{\l} and Vougioukas, Konstantinos and He, Sen and Zi{\k{e}}ba, Maciej and Petridis, Stavros and Pantic, Maja},
  booktitle={Proceedings of the IEEE/CVF Winter Conference on Applications of Computer Vision},
  pages={5091--5100},
  year={2024}
}

@inproceedings{zhu2016face,
  title={Face alignment across large poses: A 3d solution},
  author={Zhu, Xiangyu and Lei, Zhen and Liu, Xiaoming and Shi, Hailin and Li, Stan Z},
  booktitle={Proceedings of the IEEE conference on computer vision and pattern recognition},
  pages={146--155},
  year={2016}
}

@inproceedings{li2023one,
  title={One-shot high-fidelity talking-head synthesis with deformable neural radiance field},
  author={Li, Weichuang and Zhang, Longhao and Wang, Dong and Zhao, Bin and Wang, Zhigang and Chen, Mulin and Zhang, Bang and Wang, Zhongjian and Bo, Liefeng and Li, Xuelong},
  booktitle={Proceedings of the IEEE/CVF Conference on Computer Vision and Pattern Recognition},
  pages={17969--17978},
  year={2023}
}

@inproceedings{chan2022efficient,
  title={Efficient geometry-aware 3d generative adversarial networks},
  author={Chan, Eric R and Lin, Connor Z and Chan, Matthew A and Nagano, Koki and Pan, Boxiao and De Mello, Shalini and Gallo, Orazio and Guibas, Leonidas J and Tremblay, Jonathan and Khamis, Sameh and others},
  booktitle={Proceedings of the IEEE/CVF conference on computer vision and pattern recognition},
  pages={16123--16133},
  year={2022}
}

@article{li2024generalizable,
  title={Generalizable one-shot 3D neural head avatar},
  author={Li, Xueting and De Mello, Shalini and Liu, Sifei and Nagano, Koki and Iqbal, Umar and Kautz, Jan},
  journal={Advances in Neural Information Processing Systems},
  volume={36},
  year={2024}
}

@article{mildenhall2021nerf,
  title={Nerf: Representing scenes as neural radiance fields for view synthesis},
  author={Mildenhall, Ben and Srinivasan, Pratul P and Tancik, Matthew and Barron, Jonathan T and Ramamoorthi, Ravi and Ng, Ren},
  journal={Communications of the ACM},
  volume={65},
  number={1},
  pages={99--106},
  year={2021},
  publisher={ACM New York, NY, USA}
}

@inproceedings{guo2021ad,
  title={Ad-nerf: Audio driven neural radiance fields for talking head synthesis},
  author={Guo, Yudong and Chen, Keyu and Liang, Sen and Liu, Yong-Jin and Bao, Hujun and Zhang, Juyong},
  booktitle={Proceedings of the IEEE/CVF international conference on computer vision},
  pages={5784--5794},
  year={2021}
}

@inproceedings{ma2023otavatar,
  title={Otavatar: One-shot talking face avatar with controllable tri-plane rendering},
  author={Ma, Zhiyuan and Zhu, Xiangyu and Qi, Guo-Jun and Lei, Zhen and Zhang, Lei},
  booktitle={Proceedings of the IEEE/CVF Conference on Computer Vision and Pattern Recognition},
  pages={16901--16910},
  year={2023}
}

@article{rivero2024rig3dgs,
  title={Rig3dgs: Creating controllable portraits from casual monocular videos},
  author={Rivero, Alfredo and Athar, ShahRukh and Shu, Zhixin and Samaras, Dimitris},
  journal={arXiv preprint arXiv:2402.03723},
  year={2024}
}

@article{wang2023gaussianhead,
  title={Gaussianhead: Impressive head avatars with learnable gaussian diffusion},
  author={Wang, Jie and Xie, Jiu-Cheng and Li, Xianyan and Xu, Feng and Pun, Chi-Man and Gao, Hao},
  journal={arXiv preprint arXiv:2312.01632},
  year={2023}
}

@inproceedings{isola2017image,
  title={Image-to-image translation with conditional adversarial networks},
  author={Isola, Phillip and Zhu, Jun-Yan and Zhou, Tinghui and Efros, Alexei A},
  booktitle={Proceedings of the IEEE conference on computer vision and pattern recognition},
  pages={1125--1134},
  year={2017}
}

@inproceedings{thies2016face2face,
  title={Face2face: Real-time face capture and reenactment of rgb videos},
  author={Thies, Justus and Zollhofer, Michael and Stamminger, Marc and Theobalt, Christian and Nie{\ss}ner, Matthias},
  booktitle={Proceedings of the IEEE conference on computer vision and pattern recognition},
  pages={2387--2395},
  year={2016}
}

@inproceedings{wang2021one,
  title={One-shot free-view neural talking-head synthesis for video conferencing},
  author={Wang, Ting-Chun and Mallya, Arun and Liu, Ming-Yu},
  booktitle={Proceedings of the IEEE/CVF conference on computer vision and pattern recognition},
  pages={10039--10049},
  year={2021}
}

@inproceedings{zhang2021flow,
  title={Flow-guided one-shot talking face generation with a high-resolution audio-visual dataset},
  author={Zhang, Zhimeng and Li, Lincheng and Ding, Yu and Fan, Changjie},
  booktitle={Proceedings of the IEEE/CVF Conference on Computer Vision and Pattern Recognition},
  pages={3661--3670},
  year={2021}
}

@inproceedings{zhang2018unreasonable,
  title={The unreasonable effectiveness of deep features as a perceptual metric},
  author={Zhang, Richard and Isola, Phillip and Efros, Alexei A and Shechtman, Eli and Wang, Oliver},
  booktitle={Proceedings of the IEEE conference on computer vision and pattern recognition},
  pages={586--595},
  year={2018}
}

@inproceedings{ye2022perceivingli2023one,
  title={Perceiving and modeling density for image dehazing},
  author={Ye, Tian and Zhang, Yunchen and Jiang, Mingchao and Chen, Liang and Liu, Yun and Chen, Sixiang and Chen, Erkang},
  booktitle={European conference on computer vision},
  pages={130--145},
  year={2022},
  organization={Springer}
}

@article{xu2024hallo,
  title={Hallo: Hierarchical audio-driven visual synthesis for portrait image animation},
  author={Xu, Mingwang and Li, Hui and Su, Qingkun and Shang, Hanlin and Zhang, Liwei and Liu, Ce and Wang, Jingdong and Yao, Yao and Zhu, Siyu},
  journal={arXiv preprint arXiv:2406.08801},
  year={2024}
}

@article{liu2024vqtalker,
  title={VQTalker: Towards Multilingual Talking Avatars through Facial Motion Tokenization},
  author={Liu, Tao and Ma, Ziyang and Chen, Qi and Chen, Feilong and Fan, Shuai and Chen, Xie and Yu, Kai},
  journal={arXiv preprint arXiv:2412.09892},
  year={2024}
}

@article{li2017learning,
  title={Learning a model of facial shape and expression from 4D scans.},
  author={Li, Tianye and Bolkart, Timo and Black, Michael J and Li, Hao and Romero, Javier},
  journal={ACM Trans. Graph.},
  volume={36},
  number={6},
  pages={194--1},
  year={2017}
}

@article{aneja2024gaussianspeech,
  title={GaussianSpeech: Audio-Driven Gaussian Avatars},
  author={Aneja, Shivangi and Sevastopolsky, Artem and Kirschstein, Tobias and Thies, Justus and Dai, Angela and Nie{\ss}ner, Matthias},
  journal={arXiv preprint arXiv:2411.18675},
  year={2024}
}

@article{taubner2024cap4d,
  title={CAP4D: Creating Animatable 4D Portrait Avatars with Morphable Multi-View Diffusion Models},
  author={Taubner, Felix and Zhang, Ruihang and Tuli, Mathieu and Lindell, David B},
  journal={arXiv preprint arXiv:2412.12093},
  year={2024}
}

@article{kerbl20233d,
  title={3D Gaussian splatting for real-time radiance field rendering.},
  author={Kerbl, Bernhard and Kopanas, Georgios and Leimk{\"u}hler, Thomas and Drettakis, George},
  journal={ACM Trans. Graph.},
  volume={42},
  number={4},
  pages={139--1},
  year={2023}
}

@article{liu2009analysis,
  title={An analysis of the current and future state of 3D facial animation techniques and systems},
  author={Liu, Chen},
  year={2009},
  publisher={Simon Fraser University}
}

@article{saunders2024dubbing,
  title={Dubbing for Everyone: Data-Efficient Visual Dubbing using Neural Rendering Priors},
  author={Saunders, Jack and Namboodiri, Vinay},
  journal={arXiv preprint arXiv:2401.06126},
  year={2024}
}

@article{poole2022dreamfusion,
  title={Dreamfusion: Text-to-3d using 2d diffusion},
  author={Poole, Ben and Jain, Ajay and Barron, Jonathan T and Mildenhall, Ben},
  journal={arXiv preprint arXiv:2209.14988},
  year={2022}
}

@article{karras2022elucidating,
  title={Elucidating the design space of diffusion-based generative models},
  author={Karras, Tero and Aittala, Miika and Aila, Timo and Laine, Samuli},
  journal={Advances in neural information processing systems},
  volume={35},
  pages={26565--26577},
  year={2022}
}

@inproceedings{dhamo2024headgas,
  title={Headgas: Real-time animatable head avatars via 3d gaussian splatting},
  author={Dhamo, Helisa and Nie, Yinyu and Moreau, Arthur and Song, Jifei and Shaw, Richard and Zhou, Yiren and P{\'e}rez-Pellitero, Eduardo},
  booktitle={European Conference on Computer Vision},
  pages={459--476},
  year={2024},
  organization={Springer}
}

@article{watson2022novel,
  title={Novel view synthesis with diffusion models},
  author={Watson, Daniel and Chan, William and Martin-Brualla, Ricardo and Ho, Jonathan and Tagliasacchi, Andrea and Norouzi, Mohammad},
  journal={arXiv preprint arXiv:2210.04628},
  year={2022}
}

@article{tewari2023diffusion,
  title={Diffusion with forward models: Solving stochastic inverse problems without direct supervision},
  author={Tewari, Ayush and Yin, Tianwei and Cazenavette, George and Rezchikov, Semon and Tenenbaum, Josh and Durand, Fr{\'e}do and Freeman, Bill and Sitzmann, Vincent},
  journal={Advances in Neural Information Processing Systems},
  volume={36},
  pages={12349--12362},
  year={2023}
}

@article{liu2023one,
  title={One-2-3-45: Any single image to 3d mesh in 45 seconds without per-shape optimization},
  author={Liu, Minghua and Xu, Chao and Jin, Haian and Chen, Linghao and Varma T, Mukund and Xu, Zexiang and Su, Hao},
  journal={Advances in Neural Information Processing Systems},
  volume={36},
  pages={22226--22246},
  year={2023}
}

@inproceedings{muller2023diffrf,
  title={Diffrf: Rendering-guided 3d radiance field diffusion},
  author={M{\"u}ller, Norman and Siddiqui, Yawar and Porzi, Lorenzo and Bulo, Samuel Rota and Kontschieder, Peter and Nie{\ss}ner, Matthias},
  booktitle={Proceedings of the IEEE/CVF Conference on Computer Vision and Pattern Recognition},
  pages={4328--4338},
  year={2023}
}

@article{song2020denoising,
  title={Denoising diffusion implicit models},
  author={Song, Jiaming and Meng, Chenlin and Ermon, Stefano},
  journal={arXiv preprint arXiv:2010.02502},
  year={2020}
}

@inproceedings{perez2018film,
  title={Film: Visual reasoning with a general conditioning layer},
  author={Perez, Ethan and Strub, Florian and De Vries, Harm and Dumoulin, Vincent and Courville, Aaron},
  booktitle={Proceedings of the AAAI conference on artificial intelligence},
  volume={32},
  number={1},
  year={2018}
}

@inproceedings{karras2019style,
  title={A style-based generator architecture for generative adversarial networks},
  author={Karras, Tero and Laine, Samuli and Aila, Timo},
  booktitle={Proceedings of the IEEE/CVF conference on computer vision and pattern recognition},
  pages={4401--4410},
  year={2019}
}

@InProceedings{Chung16a,
  author       = "Chung, J.~S. and Zisserman, A.",
  title        = "Out of time: automated lip sync in the wild",
  booktitle    = "Workshop on Multi-view Lip-reading, ACCV",
  year         = "2016",
}

@article{conneau2020unsupervised,
  title={Unsupervised cross-lingual representation learning for speech recognition},
  author={Conneau, Alexis and Baevski, Alexei and Collobert, Ronan and Mohamed, Abdelrahman and Auli, Michael},
  journal={arXiv preprint arXiv:2006.13979},
  year={2020}
}

@inproceedings{szymanowicz2024splatter,
  title={Splatter image: Ultra-fast single-view 3d reconstruction},
  author={Szymanowicz, Stanislaw and Rupprecht, Chrisitian and Vedaldi, Andrea},
  booktitle={Proceedings of the IEEE/CVF conference on computer vision and pattern recognition},
  pages={10208--10217},
  year={2024}
}

@inproceedings{
    chu2024gagavatar,
    title={Generalizable and Animatable Gaussian Head Avatar},
    author={Xuangeng Chu and Tatsuya Harada},
    booktitle={The Thirty-eighth Annual Conference on Neural Information Processing Systems},
    year={2024},
    url={https://openreview.net/forum?id=gVM2AZ5xA6}
}

@misc{
    chu2025artalk,
    title={ARTalk: Speech-Driven 3D Head Animation via Autoregressive Model}, 
    author={Xuangeng Chu and Nabarun Goswami and Ziteng Cui and Hanqin Wang and Tatsuya Harada},
    year={2025},
    eprint={2502.20323},
    archivePrefix={arXiv},
    primaryClass={cs.CV},
    url={https://arxiv.org/abs/2502.20323}, 
}

@article{kim2024nerffacespeech,
  title={NeRFFaceSpeech: One-shot Audio-driven 3D Talking Head Synthesis via Generative Prior},
  author={Kim, Gihoon and Seo, Kwanggyoon and Cha, Sihun and Noh, Junyong},
  journal={arXiv preprint arXiv:2405.05749},
  year={2024}
}
%
% \begin{thebibliography}{8}
% \bibitem{ref_article1}
% Author, F.: Article title. Journal \textbf{2}(5), 99--110 (2016)

% \bibitem{ref_lncs1}
% Author, F., Author, S.: Title of a proceedings paper. In: Editor,
% F., Editor, S. (eds.) CONFERENCE 2016, LNCS, vol. 9999, pp. 1--13.
% Springer, Heidelberg (2016). \doi{10.10007/1234567890}

% \bibitem{ref_book1}
% Author, F., Author, S., Author, T.: Book title. 2nd edn. Publisher,
% Location (1999)

% \bibitem{ref_proc1}
% Author, A.-B.: Contribution title. In: 9th International Proceedings
% on Proceedings, pp. 1--2. Publisher, Location (2010)

% \bibitem{ref_url1}
% LNCS Homepage, \url{http://www.springer.com/lncs}, last accessed 2023/10/25
% \end{thebibliography}
\end{document}